\documentclass{article}

\usepackage{PRIMEarxiv}

\usepackage[utf8]{inputenc} 
\usepackage[T1]{fontenc}    
\usepackage{hyperref}       
\usepackage{url}            
\usepackage{booktabs}       
\usepackage{amsfonts}       
\usepackage{nicefrac}       
\usepackage{microtype}      
\usepackage{lipsum}
\usepackage{graphicx}
\graphicspath{{media/}}     

\usepackage{svg}
\usepackage{subfig,float,soul,graphicx,amsmath}
\usepackage{stfloats}

\title{2D Bidirectional Gated Recurrent Unit Convolutional Neural Networks for End-to-End Violence Detection in Videos
\thanks{\textit{\underline{Citation}}: 
\textbf{A. Traoré and M. Akhloufi, “2D Bidirectional Gated Recurrent Unit Convolutional Neural Networks for End-to-End Violence Detection in Videos,” 2020. doi: 10.1007/978-3-030-50347-5\_14.
}} 
}

\author{
  Abdarahmane Traor\'e and Moulay A. Akhloufi, \textit{Senior Member IEEE}\\
  \textit{Perception, Robotics, and Intelligent Machines Research Group (PRIME)} \\
  \textit{Department of Computer Science, Universit\'e de Moncton}\ \\
  Moncton, NB, Canada\\
  \textit{{\{eat4651, moulay.akhloufi\}@umoncton.ca}} \\
}

\begin{document}
\maketitle

\begin{abstract}
Abnormal behavior detection, action recognition, fight and violence detection in videos is an area that has attracted a lot of interest in recent years. In this work, we propose an architecture that combines a Bidirectional Gated Recurrent Unit (BiGRU) and a 2D Convolutional Neural Network (CNN) to detect violence in video sequences. A CNN is used to extract spatial characteristics from each frame, while the BiGRU extracts temporal and local motion characteristics using CNN extracted features from multiple frames. The proposed end-to-end deep learning network is tested in three public datasets with varying scene complexities. The proposed network achieves accuracies up to 98\%. The obtained results are promising and show the performance of the proposed end-to-end approach.
\end{abstract}


\section{Introduction}
Nowadays we face growing violence and criminality due to the increase in population especially in large cities \cite{mt_increasing_2016}. The surveillance systems as we know them are not reliable nor efficient because it takes a person behind the monitors to detect abnormal behavior. In the case of a grocery store we have 2 or 3 screens, but in the case of an airport, with hundreds of monitors, it is almost impossible for a human to be efficient in this security task. Over time we have seen several approaches proposed to solve this problem using computer vision. But these methods were limited by the need of a manual extraction of the characteristics and then their classification using a classifier such as SVM. With the progress in terms of hardware and software and with the rise of deep learning, new end-to-end image processing techniques have emerged. Indeed, deep architectures such as EfficientNet achieved a 97.1\% accuracy in the the challenge of classifying images among 1000 categories \cite{tan_efficientnet:_2019}. 
For videos, we are interested in action recognition. This last term defines several other sub-domains including the one studied in this article, namely violence detection. 
The main challenge of action recognition in this context is to recognize acts of violence from video images. We need to take into account two important factors: the image itself and the correlation that exists between the images over time. 
Many deep learning and vision-based techniques were proposed for violence detection. In general, they can be categorized into five type of approaches as described in the following.\newline
The first category of algorithms use 3D CNN. These algorithms are computationally expensive. However, they have proven to be efficient \cite{ding_violence_2014}, \cite{li_end_end_2018}, \cite{song_novel_2019}, \cite{Ullah2019}. 
For example,  Song \textit{et al.} \cite{song_novel_2019} propose the use of 3D CNN to enable the extraction of both spatial and temporal characteristics. The CNN is mainly composed of eight 3D convolutional layers and five 3D max pooling. 
In \cite{Ullah2019}, the authors propose a three stage deep learning approach for violence detection. They first detect persons in surveillance videos using CNN. Then, they feed a 3D CNN with a sequence of 16 frames with the detected persons. The extracted spatiotemporal features are sent to a Softmax classifier for violence detection.\newline
Other algorithms use 2D CNN in a time series approach and then refine the characteristics with a Long short-term memory (LSTM). This end-to-end method is very efficient in term of computation and gives very interesting results as reported in \cite{abdali_robust_2019}.\newline
Algorithms that use feature extraction followed by an LSTM are the most popular. They use several types of feature extraction, ranging from classical methods based on a simple CNN as in \cite{nguyen_violent_2019} to more sophisticated approaches as in \cite{ditsanthia_video_2018}.\newline
In \cite{ditsanthia_video_2018}, Ditsanthia \textit{et al.} use a method called multiscale convolutional features extraction to handle variations in the video data, and feeds the LSTMs with these multiscale extracted characteristics.\newline
Other techniques, use both optical flow and frames as input in a fusion scheme and then classify the data to detect specific actions \cite{xu_violent_2018},\cite{simonyan_two-stream_2014}.\newline
The final category uses ConvLSTMs which are LSTMs whose matrix operations have been replaced by convolutions \cite{macintyre_detecting_2019},\cite{sudhakaran_learning_2017}. Sudhakaran \textit{et al.} \cite{sudhakaran_learning_2017} use this special convolutions to classify actions from the extracted 3D data.\newline
In this work, we propose an end-to-end 2D deep learning approach to automate violence detection from video sequences. The proposed architecture outperforms many available techniques. The effectiveness of our approach is validated on three public datasets.

\section{Proposed Method}
\label{sec:method}

In this work we develop an end-to-end deep learning network which is easy to train and capable of reaching the same level of performance as existing state-of-the-art methods. 
We use a 2D CNN distributed in time to capture the temporary spatial characteristics. 
We combine it with a bidirectional GRU (BiGRU) to refine our detections.




\subsection{Convolutional Neural Network}
\label{ssec:cnn}

\begin{figure}[!tbp]
  \centering
 \includegraphics[width=8.5cm]{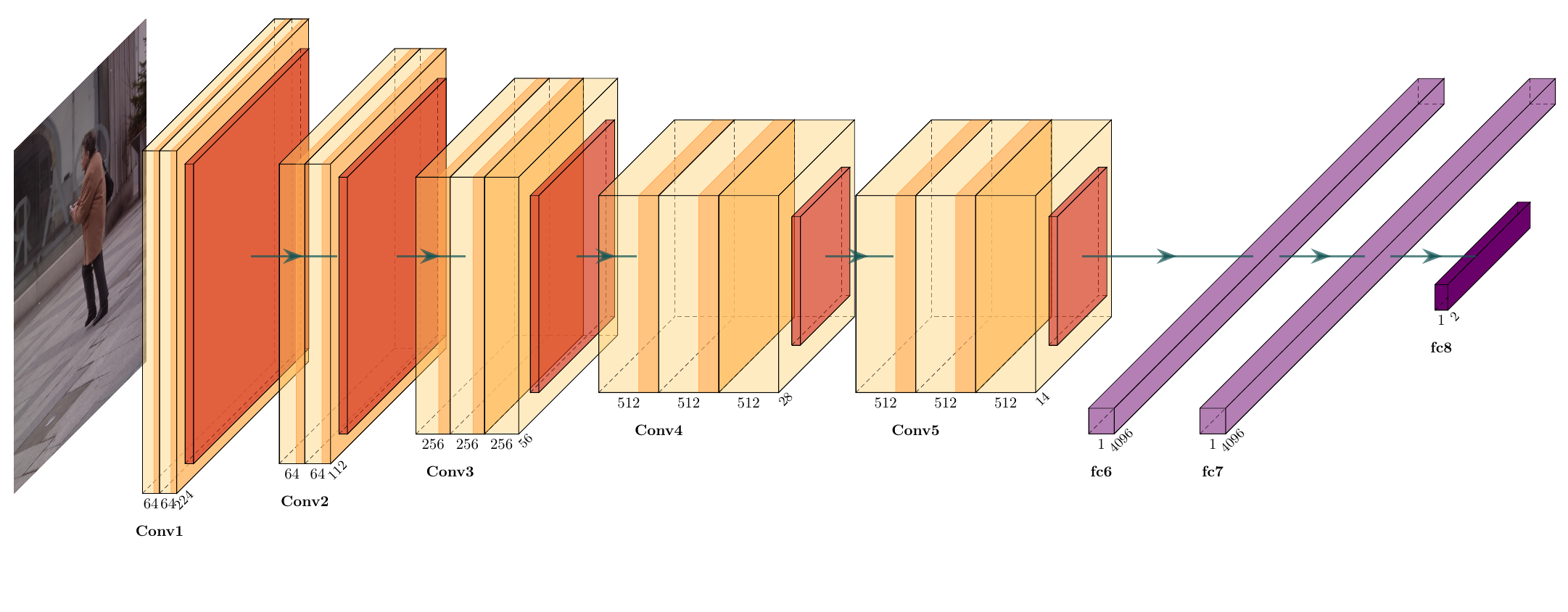}
\caption{VGG16 used for capture spatial features}
\label{figure:vgg}
\end{figure}

Our CNN network is based on a VGG16 adapted to our purpose and distributed over time to handle a squence of images. 
We pre-trained VGG on the INRA person dataset introduce by \cite{dalal_histograms_2005} which contains images of people as positives and random images as negatives. 
The objective is to identify persons in an image and ignore the background. 
After training on INRA, we remove the layers FC6, FC7, and FC8 (See figure \ref{figure:vgg}) and use a flatten layer to forward the features to the BiGRU network. 

\subsection{GATED RECURRENT UNIT}
\label{ssec:gru}

\begin{figure}[!bp]
  \centering
 \includegraphics[width=8.5cm]{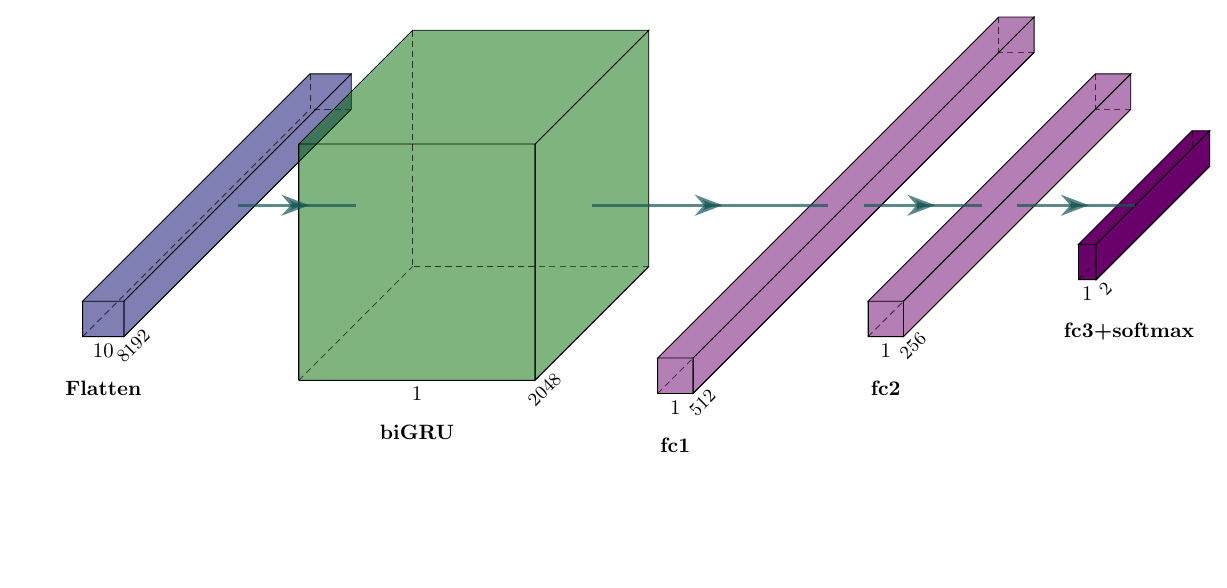}
\caption{BiGRU used to capture temporal features}
\label{fig:gru}
\end{figure}

To better consider all the spatial-temporal relationship between frames we use a GRU (See figure \ref{fig:gru}) to refine our characteristics. 
The choice of the GRU is made to avoid the gradient vanishing problem of the LSTMs. 
GRU is more robust to noise \cite{TangGRU2017} and outperforms the LSTM in several tasks \cite{KanaiGRU2017}.
Moreover, the GRUs are less computationally expensive as they have two gates unlike the LSTM which has four. 
In addition, we use our GRU in a bidirectional position to consider not only the previous sequences but also the following sequences, which allows to increase accuracy.\newline
Equation \ref{eq:t} gives the GRU functions used in this work, the update gate $Z_{t}$ decides what information to keep or to drop, the reset gate $r_{t} $  decides how much past information to forget and $h_{t}$  is the output gate. W, U and b are parameter matrices and vector, $\sigma_{g}$ is a sigmoid activation function and $\sigma_{h}$ is a hyperbolic tangent function. Finally $x_{t}$ is the input vector.

\begin{equation}
  \label{eq:t}
  \begin{gathered}
    Z_{t} = \sigma_{g}(W_{z}x_{t}+U_{z}h_{h-1}+b_{z}) \\
    r_{t} = \sigma_{g}(W_{r}x_{t}+U_{z}h_{h-1}+b_{z}) \\
    h_{t} = (1-z_{t}) \circ h_{t-1} + z_{t} \circ \sigma_{h}(W_{h}x_{t} +\\
    U_{h}(t_{t} \circ h_{t-1}) + b_{h}) 
  \end{gathered}
\end{equation}

\begin{figure}[!tbp]
    \centering
 \includegraphics[width=0.95\linewidth]{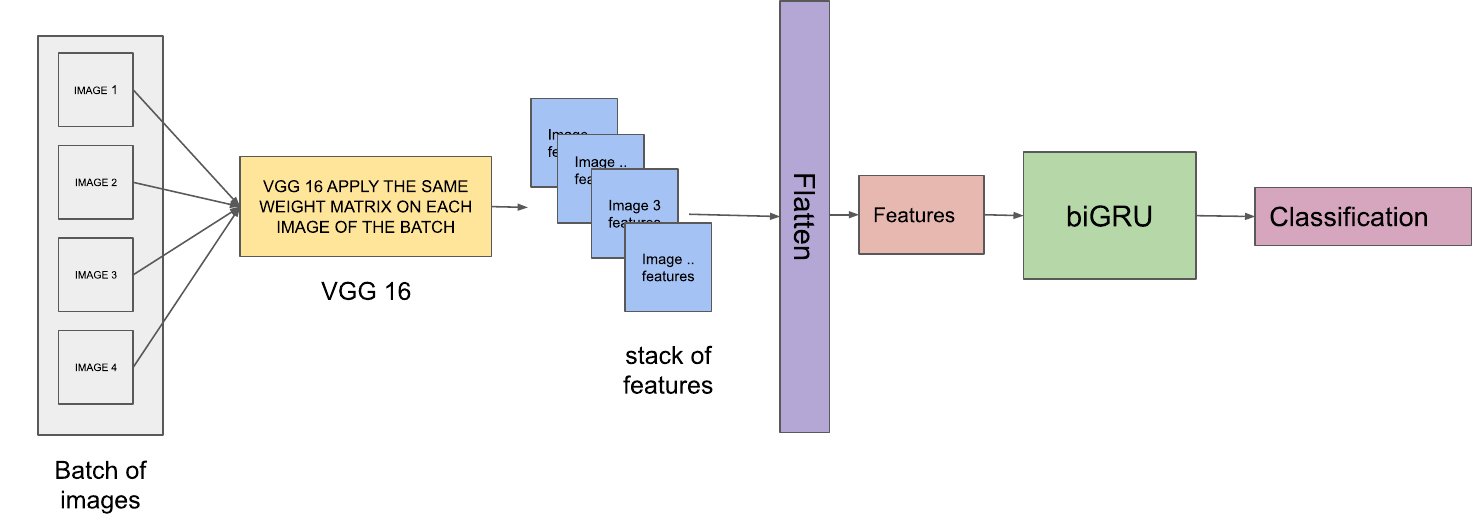}
 \caption{2D BiGRU-CNN architecture}
    \label{fig:net}
\end{figure}

The full network resulting from merging both our VGG-based CNN and BiGRU is illustrated by figure \ref{fig:net}.

\section{ Experiments and Results}
\label{sec:expares}

To evaluate our network, we used three datasets: Hockey dataset \cite{nievas2011violence}, Violent  Flow dataset \cite{hassner_violent_2012} and Real Life Violence Situations dataset \cite {noauthor_real_nodate}. We used the accuracy metric to assess the performance of our network.

\subsection{Datasets}
\label{ssec:datasets}

We divided each dataset randomly into 2 parts, training (80\%) and validation/testing (20\%). Even if with Violentflow we do not have enough data, this will allow us to demonstrate the network generalization capacity on a small dataset.

\subsubsection{Hockey Dataset}
\label{sssec:Hockey}

\begin{figure}[!tbp]
  \centering
  \subfloat{\includegraphics[width=0.32\textwidth, height=0.3\textwidth]{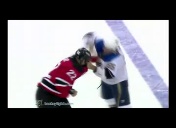}} 
  \hfill
  \subfloat{\includegraphics[width=0.32\textwidth, height=0.3\textwidth]{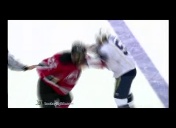}}
    \hfill
  \subfloat{\includegraphics[width=0.32\textwidth, height=0.3\textwidth]{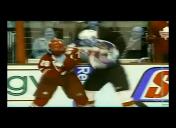}}
  \caption{Frames from Hockey Dataset}
  \label{fig:hockey}
\end{figure}

Hockey fight dataset \cite{nievas2011violence}, is a dataset of 1000 fight and no fight sequences of hockey matches (see figure \ref{fig:hockey}). 
In this dataset, the classes are balanced. Each clip lasts approximately 2 seconds and consists of about 41 frames with a resolution of 360x288. 
The details captured by the the sequences are quite similar, especially the background. 
We resized the images to 128x176. 

\subsubsection{Violent Flow Dataset}
\label{sssec:vioflow}

\begin{figure}[!tbp]

  \centering
  \subfloat{\includegraphics[width=0.32\textwidth]{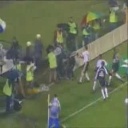}}
  \hfill
  \subfloat{\includegraphics[width=0.32\textwidth]{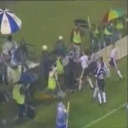}}
    \hfill
  \subfloat{\includegraphics[width=0.32\textwidth]{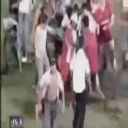}}
  \caption{Frames from ViolentFlow}
    \label{fig:flow}
  
\end{figure}

Violent Flow Dataset \cite{hassner_violent_2012} is a dataset of real-world video footage of crowd violence (see figure \ref{fig:flow}). 
The dataset contains 246 videos. 
The shortest clip duration is 1.04 seconds, the longest clip is 6.52 seconds, and the average length of a video clip is 3.60 seconds. 
We also resized the images to 128x128. 

\subsubsection{Real Life Violence Situations Dataset}
\label{sssec:rlviolence}

\begin{figure}[!tbp]
\vspace{-0.2cm}

  \centering
  \subfloat{\includegraphics[width=0.32\textwidth]{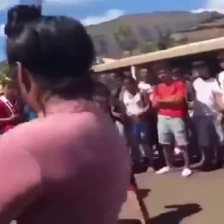}}
   \hfill
  \subfloat{\includegraphics[width=0.32\textwidth]{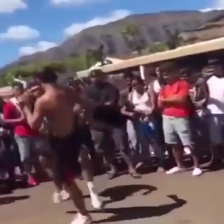}}
     \hfill
   \subfloat{\includegraphics[width=0.32\textwidth]{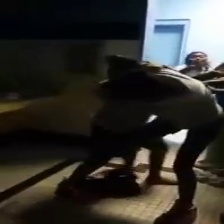}}
  
  \caption{Frames from Real Life Violence Situations Dataset}
  \label{fig:real}
\end{figure}

Real-Life Violence Situations Dataset \cite{noauthor_real_nodate} is a dataset of violence video clips (see figure \ref{fig:real}). It contains 1000 fight and no fight sequences (classes are balanced). 
The dataset is homogeneous, with various types of situations.
We resized the images to 128x128. 

\subsection{Parameters and sampling}
\label{ssec:settings}

For the implementation of our network, we used Keras with the TensorFlow backend. 
Since we do not have the same video lengths and the videos are at average 30 fps, we have chosen a different number of frames for each dataset and we used a uniform sampling method. \newline
With this sampling, not only the number of parameters to compute decreases, but we avoid unnecessary calculations of redundant frames. \newline
Before starting the training on the datasets, we pretrained the VGG-based CNN on INRA for 100 epochs using SGD, a learning rate of 0.001 and a batch size of 8. 
Then, we combined our 2D CNN with BiGRU followed by 3 fully connected layers.\newline
The proposed 2D BiGRU-CNN was trained on all the datasets using SGD as optimizer with a batch size of 10. The learning rate was set to 0.0008 for the Hockey and Real Life Violence datasets. The best learning rate for Violent Flow was 0.0006.
The network was trained for 250 epochs.
We used 10 frames of each video as the input to our network.




\subsection{Results and discussions}
\label{ssec:result}

In table \ref{table:hockey} and \ref{table:flow} we present an accuracy comparison for the two datasets Hockey and Violent Flow with previously published techniques.\newline
On the Hockey dataset we obtained an accuracy of 98\%, and on Violent Flow we obtained an accuracy of 95.5\%. 
We can see from tables \ref{table:hockey} and \ref{table:flow} that our approach outperforms most of the previously published techniques. 
It is only surpassed by the use of 3D CNNs in \cite{song_novel_2019} for Hockey and \cite{Ullah2019} for Violent Flow. 
Still we obtain better results in Hockey compared to the work in \cite{Ullah2019} and in Violent Flow compared to the model proposed in \cite{song_novel_2019}. 
These last techniques use computationally intensive 3D convolutions \cite{kanojia2019exploring}, whereas we only use 2D CNN. \newline
Finally, Real-Life Violence Situations is the dataset that allows us to confirm that the proposed network is robust, because this last dataset contains various situations (We find Hockey scenes, outdoor scenes, etc.). 
With the obtained accuracy of 90.25\%, we can say that our network has a good generalization. \newline
In term of computation complexity, our approach depends more on CNN than on GRU. 
Indeed, in our case we use VGG16 having 138 million parameters. 
We use it with smaller image sizes, which reduces the number of parameters.
Moreover the use of GRU helps reduce the overall complexity of the architecture.




\begin{table}[h!t]
\setlength{\tabcolsep}{3.7em}
\caption{Hockey Dataset comparison with other methods}
\label{table:hockey}
\begin{center}
 \begin{tabular}{||c c ||} 
 \hline
 Method & Accuracy (\%) \\ [0.5ex] 
 \hline\hline
 HOF + BoW & 88.6 \cite{real_violence_2011}\\ 
 \hline
 HOG + BoW  & 91.7  \cite{real_violence_2011} \\
 \hline
 MoSIFT + BoW  & 90.9 \cite{real_violence_2011} \\
 \hline
  MoWLD + BoW  & 91.9  \cite{zhang_mowld:_2017}\\
 \hline
 MoWLD + Sparse Coding  & 93.7 \cite{zhang_mowld:_2017}\  \\
 \hline
 MoSIFT + KDE + Sparce Coding  & 94.3  \cite{xu_violent_2014} \\
 \hline
 MoWLD + KDE + Sparce Coding  &   94.9 \cite{zhang_mowld:_2017}\\
 \hline
 MoWLD + KDE +SRC  &  96.8 \cite{song_novel_2019} \\
 \hline
  3D-CNN  &   91 \cite{song_novel_2019}\\
 \hline
 3D ConvNet (16 frames) &   96 \cite{Ullah2019}\\
 \hline
  FightNet &   97 \cite{fnet}\\
 \hline
   VGG19+LSTM &   97 \cite{abdali_robust_2019}\\
 \hline
   ConvLSTM & 97.1 \cite{sudhakaran_learning_2017}\\
 \hline
   \textbf{Our approach} &   \textbf{98}\\
 \hline
  3D ConvNet (16 frames) &  98.96 \cite{song_novel_2019}\\
 \hline
  3D ConvNet (32 frames) &   99.62 \cite{song_novel_2019}\\
 \hline
\end{tabular}

\end{center}
\end{table}

\begin{table}[h!t]
\setlength{\tabcolsep}{3em}
\caption{ViolentFlow Dataset comparison with other methods}
\label{table:flow}
\begin{center}
 \begin{tabular}{||c c ||} 
 \hline
 Method & Accuracy (\%)  \\ [0.5ex] 
 \hline\hline
 LTP & 71.53  \cite{hassner_violent_2012} \\ 
 \hline
 ViF  & 81.3 \cite{hassner_violent_2012}\\
 \hline
 MoWLD + BoW  & 82.56 \cite{zhang_mowld:_2017}\\
 \hline
 RVD  & 82.79 \cite{zhang_new_2016}\\
 \hline
 MoWLD + Sparse Coding  & 82.39   \cite{zhang_mowld:_2017}\\
 \hline
 VGG19+LSTM &   85.71 \cite{abdali_robust_2019}\\
 \hline
 MoSIFT + KDE + Sparce Coding  & 89.05  \cite{xu_violent_2014} \\
 \hline
 MoWLD + KDE + Sparce Coding  &   89.78  \cite{zhang_mowld:_2017}\\
 \hline
 MoWLD + KDE +SRC  &   93.19 \cite{song_novel_2019} \\
 \hline
 
  3D ConvNet (Unifrom sampling method) &  93.5 \cite{song_novel_2019}\\
 \hline
   ConvLSTM & 94.57 \cite{sudhakaran_learning_2017}\\
 \hline
  3D ConvNet (New Sampling method) &   94.5 \cite{song_novel_2019}\\
 \hline
 \textbf{Our approach} &   \textbf{95.5}\\
 \hline
 3D ConvNet (16 frames) & 98 \cite{Ullah2019}\\
 \hline
\end{tabular}

\end{center}
\end{table}

\newpage
\section{Conclusion and Future Work}

In this work we proposed a simple end-to-end deep learning approach to detect violence in video sequences. 
The new architecture combines 2D CNN and bidirectional GRU. 
The CNN uses a modified VGG16 pretrained on a person dataset to learn how to extract persons from video images. 
The CNN features from multiple images are sent to the bidirectional GRU which takes into account temporal and local motion characteristics. 
Tests were performed on 3 public datasets. 
The obtained results show that the proposed approach outperforms many previous approaches. 
It is only slightly surpassed by 3D CNN models. 
However, since we used 2D CNN our approach is less computationally intensive. 
Future work includes the use of different sampling approaches that have been shown to increase the performance in video classification \cite{song_novel_2019}. In addition, we will explore the fusion of frames with optical flow \cite{xu_violent_2018} in order to improve accuracy. Finally, as our approach is modular, we can get closer to real-time performance using light CNNs such as MobileNets \cite{Sandler_2018_CVPR}.

\section*{Acknowledgments}
Thanks to the Natural Sciences and Engineering Research Council of Canada (NSERC), [funding reference number RGPIN-2018-06233

\bibliographystyle{unsrt}  
\bibliography{references}

\end{document}